\documentclass[letterpaper,twocolumn,10pt]{article}
\usepackage{style/usenix}

\usepackage{color}
\usepackage{multirow}
\usepackage{array}
\usepackage{url}
\usepackage{algorithmicx,algorithm}
\usepackage[noend]{algpseudocode}
\usepackage{enumitem}
\usepackage{times}
\usepackage{xspace}
\usepackage{latexsym}
\usepackage{amsmath}
\usepackage{booktabs}
\usepackage{tabularx}
\usepackage{balance}
\usepackage{graphicx}
\usepackage{svg}
\usepackage{pifont}
\svgsetup{
    inkscapepath=i/svg-inkscape/
}
\svgpath{{svg/}}
\usepackage{caption}
\usepackage{subcaption}
\usepackage{amssymb}
\usepackage{etoolbox}
\usepackage[all]{nowidow}
\usepackage{authblk}
\usepackage{enumitem}
\setitemize{noitemsep,topsep=0pt,parsep=0pt,partopsep=0pt}
\usepackage{multirow}
\usepackage{array}
\usepackage{dblfloatfix}

\setlist{nolistsep}
\newcommand{\parabf}[1]{\medskip\noindent\textbf{#1}}
\newcommand{\parait}[1]{\medskip\noindent\textit{#1}}
\newcommand{\paraf}[1]{\noindent\textbf{#1}}
\newcommand{\cut}[1]{}

\newcommand{\sysname}{RLHFuse\xspace}

\newcommand{\revise}[1]{\textcolor{blue}{#1}}

\begin{document}
\sloppy
\date{}

\definecolor{ao}{rgb}{0.0, 0.5, 0.0}
\definecolor{method}{rgb}{0., 0., 0.}

\title{Optimizing RLHF Training for Large Language Models with Stage Fusion}

\author{
    \rm{
        Yinmin Zhong$^{\text{1}}$ \enskip
        Zili Zhang$^{\text{1}}$ \enskip
        Bingyang Wu$^{\text{1}}$ \enskip
        Shengyu Liu$^{\text{1}}$ \enskip
        Yukun Chen$^{\text{2}}$ \enskip
    }
    \vspace{-0.1in}
    \\
    \rm{
        Changyi Wan$^{\text{2}}$ \enskip
        Hanpeng Hu$^{\text{2}}$ \enskip
        Lei Xia$^{\text{2}}$ \enskip
        Ranchen Ming$^{\text{2}}$ \enskip
        Yibo Zhu$^{\text{2}}$ \enskip
        Xin Jin$^{\text{1}}$ \enskip
    }
    \\
    \vspace{0.1in}
    {$^{\text{1}}$\textit{School of Computer Science, Peking University}\enskip $^{\text{2}}$\textit{StepFun}\enskip}
}

\maketitle
\pagestyle{empty}
\captionsetup[figure]{font=small}
\captionsetup[table]{font=small}

\begin{abstract}
We present \sysname, an efficient training system with \emph{stage fusion} for
Reinforcement Learning from Human Feedback (RLHF).
Due to the intrinsic nature of RLHF training, \textit{i.e.}, the
data skewness in the generation stage and the pipeline bubbles in the training
stage, existing RLHF systems suffer from low GPU utilization.
\sysname breaks the traditional view of RLHF workflow as a
composition of individual tasks, splitting each task into finer-grained
subtasks, and performing \textit{stage fusion} to improve GPU utilization.
\sysname contains two key ideas. First, for generation and inference tasks,
\sysname splits them into sample-level subtasks, enabling efficient
\textit{inter-stage fusion} to overlap the execution of generation
and inference stages, thus mitigating the original generation bottleneck
dominated by long-tailed samples. Second, for training tasks, \sysname breaks
them into subtasks of micro-batches and
performs \textit{intra-stage fusion} to concurrently execute these subtasks in
the training stage with a fused pipeline schedule, effectively mitigating the pipeline
bubbles. The experiments show that
\sysname increases the training throughput by up to $3.7\times$, compared to
existing systems.
\end{abstract}

\vspace{-0.1in}
\section{Introduction}
\label{sec:introduction}

\begin{figure}[t!]
    \centering
    \includegraphics[width=\linewidth]{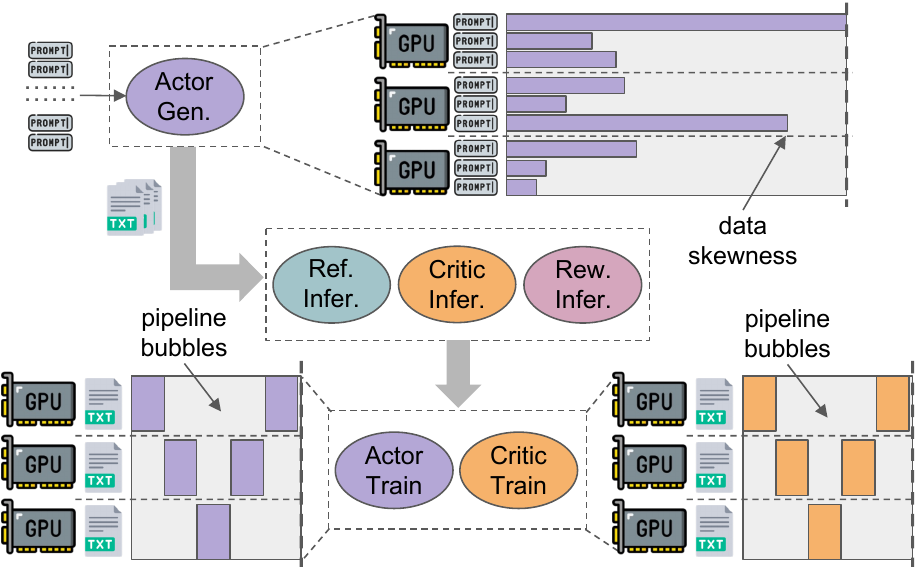}
    \vspace*{-0.2in}
    \caption{RLHF workflow~\cite{ouyang2022traininglanguagemodelsfollow} and the problems in the generation and training stages of existing RLHF training frameworks.}
    \label{fig:intro:teaser}
    \vspace*{-0.05in}
\end{figure}

The rise of Large Language Models (LLMs) marks a revolutionary leap in 
generative AI.
Despite the impressive capabilities of LLMs, many studies~\cite{gehman2020realtoxicitypromptsevaluatingneuraltoxic,llmdanger,lin2022truthfulqameasuringmodelsmimic} have shown that LLMs often display unintended behaviors.
To address these issues, 
Reinforcement Learning from Human Feedback (RLHF)~\cite{ouyang2022traininglanguagemodelsfollow} has been introduced, 
aiming to align LLMs with human intent after pre-training. 
With a concrete and well-defined reward target,
RLHF enhances the ability of LLMs in multiple domains.
Figure~\ref{fig:intro:teaser} shows the typical RLHF training workflow, which involves
six different tasks divided into three stages. 
The \texttt{Actor} model first generates samples with the input prompts and passes them to the 
\texttt{Reference}, \texttt{Critic}, and \texttt{Reward} models for inference,
finally the loss is calculated to train both the \texttt{Actor} and \texttt{Critic} models, while
the other two models remain frozen.
There are barriers between stages due to the data and weight dependencies.

To enable efficient training of RLHF, existing frameworks
~\cite{sheng2024hybridflow,mei2024realhf,lei2024puzzle}  have proposed a 
series of optimizations, such as selecting tailored parallel
strategies for each task
and optimizing the data exchange during stage transitions.
However, these techniques \textit{fundamentally treat the RLHF workflow as a 
simple composition of individual tasks}, failing to delve into the inherent 
characteristics and structure inside the tasks, thereby missing 
significant optimization opportunities. In production deployments, we have observed two issues 
within current frameworks that severely impact training efficiency,
as indicated by the grey area in Figure~\ref{fig:intro:teaser}. 

The first issue arises from the \textit{data skewness} in the generation stage, 
where the output length of generated samples follows a long-tailed distribution.
This phenomenon is broadly observed during LLM generation (\S\ref{sec:motivation:problems})
and this data diversity is critical to the robustness of RL training.
On the one hand, during the latter part of the generation stage, only a small number of long-tailed 
samples remain processing, but a large batch size is required for generation to improve 
GPU utilization (\S\ref{sec:motivation:background}). On the other hand, later tasks cannot start in advance due to data dependencies.
As a result, the generation time is dominated by the longest sample and the GPU utilization is low, 
as shown in the upper right of Figure~\ref{fig:intro:teaser}. 
Although numerous optimization techniques~\cite{vllm,yu2022orca,zheng2024sglangefficientexecutionstructured,dao2022flashattention,hong2023flashdecoding++}
are proposed to improve 
the generation speed and throughput, they cannot alter the inherent long-tail phenomenon and thus fail
to solve the problem.
Making the problem even worse, as the maximum output length of LLMs
continues to grow~\cite{gpt4olongoutput,openaio1}, this long-tail phenomenon will become more significant.

The second issue stems from the \textit{pipeline bubbles} in the training stage. 
With the explosive growth in the size of LLMs, higher pipeline parallelism (PP) size is needed to scale 
training. When LLMs reach hundreds of billions of parameters, 
it is quite common for PP size to reach a few dozen~\cite{shoeybi2020megatronlm,jiang2024megascale}. 
However, the proportion of pipeline bubbles increases with PP size, significantly reducing 
training efficiency. In RLHF, there are two training tasks, each may evolve a specific LLM
with hundreds of billions of parameters, which amplifies this inefficiency.
Despite that various approaches have been proposed
to reduce pipeline bubbles when training individual models~\cite{huang2019gpipe,pipedream,li2021chimera,qi2023zero},
the remaining bubbles still pose a significant challenge under synchronous training semantics.

To this end, we propose \sysname, which tackles the \textit{long-tailed
generation issue} and \textit{pipeline bubbles} by employing \textit{subtask-level 
optimizations}.
By breaking down the task into finer-grained subtasks,
\sysname opens up new design space to \textit{fuse the stage execution}.
This is orthogonal to current RLHF frameworks\cite{sheng2024hybridflow,
mei2024realhf, lei2024puzzle} that optimizes on the task-level to choose the optimal parallel strategies 
for each task or reduce the task switching overhead.
\sysname focuses on how to execute the RLHF workflow 
efficiently with stage fusion after the parallel strategies are chosen.
\sysname exploits the opportunities at the subtask level
and proposes two new techniques, i.e.,
\textit{data-aware inter-stage fusion} and \textit{model-aware 
intra-stage fusion}. These two techniques leverage the unique system characteristics 
of RLHF training to fuse the execution across different tasks, thus effectively addressing the data skewness and pipeline bubbles, respectively.

For data skewness, we split the generation and inference tasks into
sample-level subtasks and then the dependency granularity between the 
generation and inference stage can be refined from stage-level to sample-level. 
This is due to the fact that the computation of the two stages is 
essentially independent for different samples. To leverage this insight, 
we design a lightweight and efficient migration algorithm to automatically 
detect when the generation stage enters the inefficient long-tail processing phase 
and migrate the remaining long-tailed samples to a limited number of GPUs.
The freed-up resources are used to launch the inference tasks.
This approach enables inter-stage fusion and dynamically adjusts the migration 
timing in each iteration according to the workload, thereby maximizing overall efficiency.

For pipeline bubbles, we break the training task into subtasks of micro-batches.
Pipeline bubbles are essentially caused by dependencies between subtasks
of the same model. Fortunately, in the RLHF training stage,
there are two independent models, and their subtasks are mutually independent.
Based on this observation, we employ a fused pipeline schedule to
enable intra-stage fusion, which executes the training tasks
on the same set of GPUs with bidirectional pipelines, effectively filling each other's 
pipeline bubbles. We design a lightweight schedule generation 
algorithm that can produce near-optimal schedules for \texttt{Actor} and 
\texttt{Critic} models of any size and parallel configurations.

In addition, we apply a series of system optimizations covering each RLHF stage, transforming \sysname into a 
production-ready framework to support the RLHF training of our internal products.
We evaluate \sysname on various LLMs and real-world datasets. Compared to state-of-the-art
solutions, \sysname improves the throughput up to $3.7\times$.

In summary, we make the following contributions:
\begin{itemize}[leftmargin=*]
    \item We identify the key issues within current RLHF training frameworks and propose
    to view the RLHF workflow from a finer-grained subtask-level perspective.
    \item We present \sysname, a highly efficient RLHF training framework that utilizes
    inter- and intra-stage fusion to improve the training throughput.
    \item We conduct a comprehensive evaluation of \sysname and compare it
    with state-of-the-art RLHF training solutions.
\end{itemize} 

\vspace{-0.1in}
\section{Motivation}
In this section, we first introduce the basics of RLHF training (\S\ref{sec:motivation:background}) and then 
point out the problems in existing RLHF training systems (\S\ref{sec:motivation:problems}).

\vspace{-0.1in}
\subsection{Background}
\label{sec:motivation:background}

One complete RLHF process consists of three key steps: supervised fine-tuning,
reward model training, and model alignment using the PPO algorithm~\cite{schulman2017proximalpolicyoptimizationalgorithms}.
The first two stages both train a single LLM, which has been studied by previous
work like Megatron-LM~\cite{shoeybi2020megatronlm} and MegaScale~\cite{jiang2024megascale}. The third stage, which this
paper focuses on, is unique and complex in terms that it involves multiple
models and stages. Next, we provide a detailed overview of the RLHF models and
workflow.

\parabf{RLHF models.} 
Typically, RLHF training involves four LLMs in total: \texttt{Actor}, 
\texttt{Ref}, \texttt{Critic}, and \texttt{RW}. The \texttt{Actor} model, which is the primary 
model to be trained and the final product after RLHF, serves as the \textit{agent} in the RL semantics.
Given a prompt, each token it generates autoregressively is considered an \textit{action},
and the goal of the PPO algorithm is to guide it to produce actions that align with the reward target.
The Reference model (\texttt{Ref}) is initialized from the original \texttt{Actor} model, but is frozen (\textit{i.e.}, weights do not update) during training.
It provides Kullback-Leibler (KL) divergence regularization to ensure the \texttt{Actor} model does not
deviate excessively from its original version during training or generate nonsensical outputs. 
The Reward model (\texttt{RW}) is trained on human-labeled preference data to score each sample
generated by the \texttt{Actor} model, which guides the \texttt{Actor} model to generate responses that align with human preferences.
It is also frozen during the RLHF training. The \texttt{Critic} model, 
initialized from the \texttt{RW} model, serves as the value model to evaluate the actions taken by the \texttt{Actor}.
It provides finer-grained feedback on the action level to guide \texttt{Actor} towards better decisions.

\parabf{RLHF workflow.}
The workflow of one RLHF training iteration contains three main stages:

\parait{\underline{\textit{Generation} stage.}}
The \texttt{Actor} model generates responses for the prompts in the 
current batch. This process involves a prefill phase and a decoding phase. The prefill phase 
processes the prompt to generate the first output token.
Then the decoding phase sequentially generates subsequent tokens in multiple 
steps autoregressively (i.e., each decoding step generates a new token based on tokens generated in previous 
steps until reaching a termination token). Ultimately, each prompt and its corresponding response form 
one training sample for this iteration, referred to as a \textit{rollout} or \textit{trajectory} 
in the context of RL. To increase sample efficiency, it is common to generate 
many samples in one iteration.

\parait{\underline{\textit{Inference} stage.}} 
The \texttt{Ref}, \texttt{RW}, and \texttt{Critic} models each perform 
a forward pass on the generated samples. It is similar to the prefill phase in the generation stage
but does not generate the next token. Instead, the output logits are used to calculate the
training loss in the \textit{Training} stage. Note that the execution time of each inference 
task depends on the model size and is generally not the same.

\parait{\underline{\textit{Training} stage.}}
In the PPO algorithm, unlike traditional LLM training, all the samples are divided into several mini-batches and the model weights need to be updated after training on each mini-batch. Concretely, for each 
mini-batch, the \texttt{Actor} and \texttt{Critic} models first perform a forward pass, then calculate 
the loss using the results from the inference stage and performing a backward pass to update the parameters. 
After the \texttt{Actor} completes training on all mini-batches, it uses the \textit{up-to-date} 
parameters for the generation in the next iteration.

\parabf{LLM parallelization.}
Various parallelism methods are proposed to scale training, guided by the scaling law~\cite{kaplan2020scaling}.
Data Parallelism (DP) replicates the model weights and distributes the
data among the replicas to execute in parallel.
It requires gradient synchronization across the replicas
after each iteration.
Tensor Parallelism (TP) assigns individual operators over
multiple GPUs, with each executing part of the
computation in parallel.
It requires significant communication, so it is typically used
within a node to leverage the high-bandwidth NVLINK connections.
Pipeline Parallelism (PP) organizes LLM layers into stages,
each running on a separate device or node.
It partitions the input batch into multiple “micro-batches”
to form pipeline execution and accumulate the gradients of the entire input batch. 
Each parallel method has its own
advantages, so in practice all three parallel strategies are used together to scale
training, such as Megatron-LM~\cite{shoeybi2020megatronlm} and
MegaScale~\cite{jiang2024megascale} in pre-training~\cite{shoeybi2020megatronlm}
as well as ReaLHF~\cite{mei2024realhf} and HybridFlow~\cite{sheng2024hybridflow} in RLHF training.

\vspace{-0.08in}
\subsection{Problems in RLHF Training}
\label{sec:motivation:problems}

During our practical deployment of RLHF training, we identified two key issues in the \textit{Generation}
and \textit{Training} stage within existing RLHF training systems that contribute to substantial GPU under-utilization.

\parabf{Generation Stage: data skewness.} One major problem in the generation stage is that the response length of generated samples
exhibits a long-tailed distribution, meaning that a few samples are significantly longer than the others.
Due to data dependencies, the inference tasks cannot start until the generation task finishes. 
Consequently, even only a few long-tailed samples force the inference tasks to wait.
This leads to significantly low GPU utilization in the generation stage
because the decoding phase of LLM generation is memory-bandwidth-bound, requiring a large 
batch size (typically a few hundred) to maintain high GPU utilization.

\begin{figure}[t!]
    \centering
    \includegraphics[width=\linewidth]{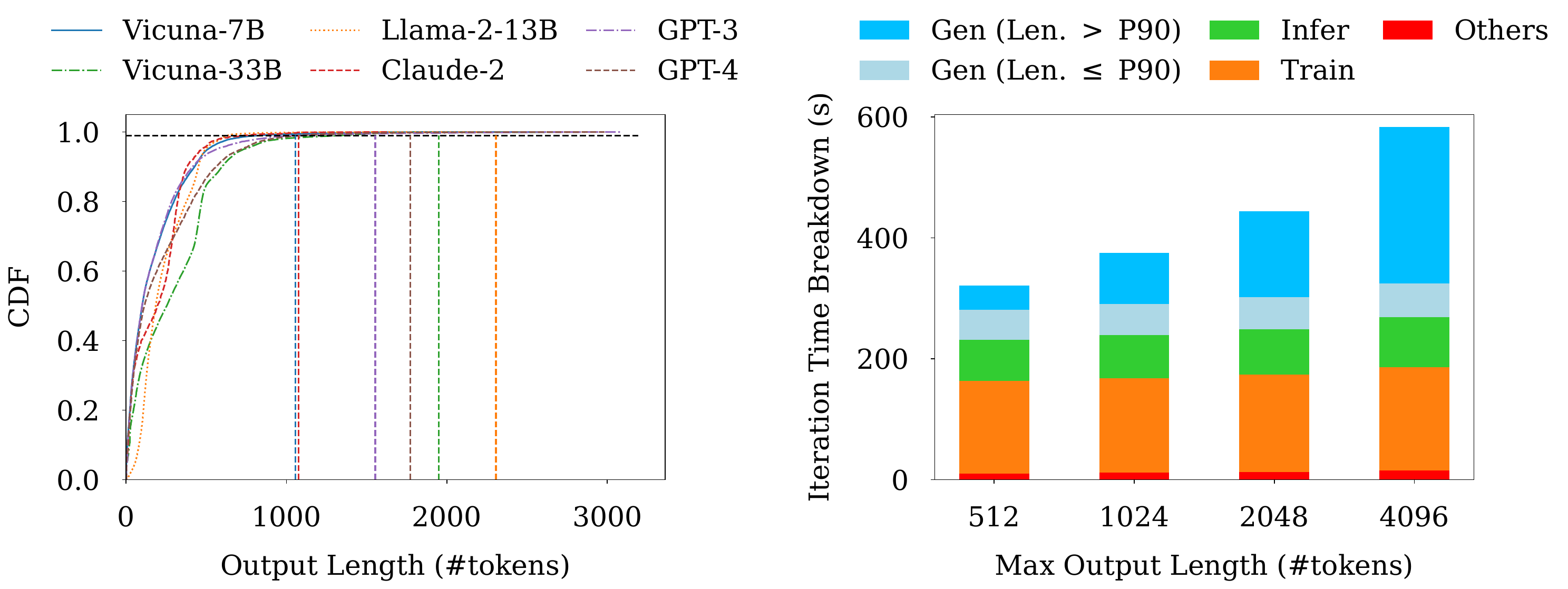}
    \vspace*{-0.25in}
    \caption{Left: The output length CDF of models in the LMSYS-Chat-1M
    dataset. The vertical dotted line indicates the P99.9 output length.
    Right: The RLHF training iteration breakdown on the internal
    model and datasets under different maximum output lengths.}
    \label{fig:motivation:long_tail}
\end{figure}

This long-tail phenomenon is prevalent and pronounced in the LLM generation. Figure~\ref{fig:motivation:long_tail} (left) shows the 
CDF of output length distribution in LMSYS-Chat-1M dataset~\cite{zheng2024lmsyschat1mlargescalerealworldllm}, which collects one million user requests and 
corresponding responses from various models on the popular LLM evaluation platform Chatbot Arena. 
We observe this long-tail distribution across models of varying sizes, 
both open-source and proprietary.
The vertical dotted lines mark the 99.9-th percentile length corresponding to each model, 
which is more than ten times the median length.
In practice, it is common to generate over 1,000 samples in each iteration, 
meaning that almost every iteration will encounter long-tailed samples.

Internally, we also observe this pattern during the RLHF training of a proprietary model with 
hundreds of billions of parameters. 
Figure~\ref{fig:motivation:long_tail} (right) presents 
the iteration time breakdown under different maximum output length settings. The time spent 
generating long-tailed samples (dark blue bar) accounts for more than half of the total generation time.
The situation gets worse as the maximum generation length increases, which leads to a 
substantial rise in iteration time. It is worth noting that this change
only affects the long-tailed samples ($< 1\%$),
but the results indicate that this very small portion of long-tailed samples can have 
a significant impact on the overall system performance. As user applications increasingly 
demand lengthy outputs~\cite{gpt4olongoutput,openaio1,deepseekr1}, the inclusion of samples with long response lengths in RLHF 
training becomes more common and important, making the long-tail phenomenon an urgent problem 
in real deployment. 

\parabf{Training Stage: pipeline bubbles.} With the exponential growth in the size of LLMs,
higher parallelism is required to scale training.
While TP size is generally limited to the number of GPUs within a single node (typically 8)
due to the significant communication overhead,
and DP size increases the memory consumption of model weights linearly.
Consequently, PP has become a critical method for scale training.

\begin{figure}[t!]
    \centering
    \includegraphics[width=\linewidth]{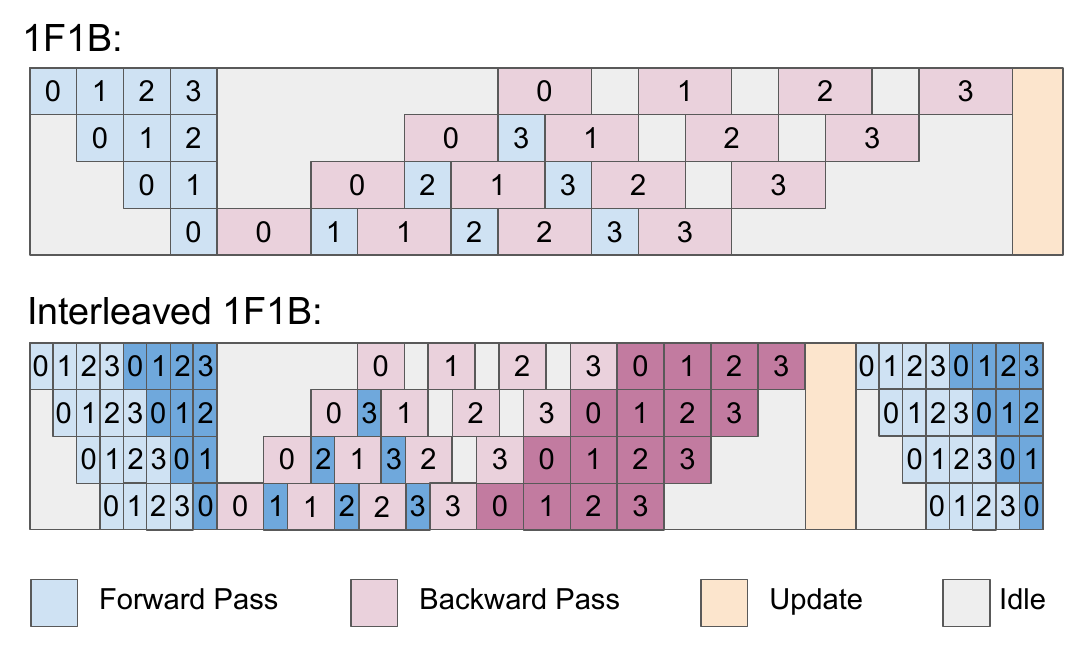}
    \vspace*{-0.3in}
    \caption{The timeline of 1F1B~\cite{pipedream} and interleaved 1F1B~\cite{shoeybi2020megatronlm} pipeline schedule with 4 pipeline stages
    and 4 micro-batches.}
    \label{fig:motivation:pp_schedule}
\end{figure}

Scaling the PP size is not a free lunch, as pipeline bubbles can significantly impact training 
efficiency. 
Concretely, in the most commonly used 1F1B pipeline schedule~\cite{pipedream} 
as shown in Figure~\ref{fig:motivation:pp_schedule} (upper), 
the \textit{pipeline bubble percentage} is
    $\frac{N - 1}{N - 1 + M}$
with $N$ PP stages and $M$ micro-batches.
One simple way to reduce the bubble 
percentage is to increase $M$. However, the global batch size 
is constrained by the converging conditions and cannot be increased indefinitely.
Additionally, in RLHF, the global batch is first divided into several mini-batches,
then distributed among the DP groups, and finally split into micro-batches, further limiting $M$.
As $N$ scales with the model size to approach $M$, which is quite common when 
scaling LLMs to hundreds of billions of parameters, the bubble percentage is about 50\%. 
This means that about half of the GPUs are idle during training, 
leading to a significant waste of resources.

Interleaved 1F1B scheduling~\cite{shoeybi2020megatronlm} is proposed to mitigate the pipeline bubbles. 
As shown in Figure~\ref{fig:motivation:pp_schedule} (bottom), it divides the LLM
into more fine-grained model chunks, with each stage hosting 
$K (K > 1)$ model chunks, reducing the pipeline bubble percentage to 
    $\frac{N - 1}{N - 1 + KM}$
.
However, it not only introduces a $K$-fold communication overhead but also 
exacerbates the imbalances across PP stages. As a result, $K$ is typically set
to a small constant in practice, leaving room for further optimizations.

Other works~\cite{pipedream, qi2023zero} aim to eliminate pipeline bubbles; however, they either 
violate the synchronous training semantics~\cite{pipedream} or rely on specific assumptions~\cite{qi2023zero}. As a result, pipeline bubbles still pose a significant problem 
in LLM training, especially when the model size keeps scaling.

\parabf{Summary.} Essentially, the above issues are caused by the inherent 
characteristics of RLHF, which are difficult to eliminate
through task-level optimizations. Existing systems
treat the task as the smallest execution unit and overlook
its internal structure, thereby missing a huge
optimization opportunities.

\begin{figure}[t!]
    \centering
    \includegraphics[width=0.9\linewidth]{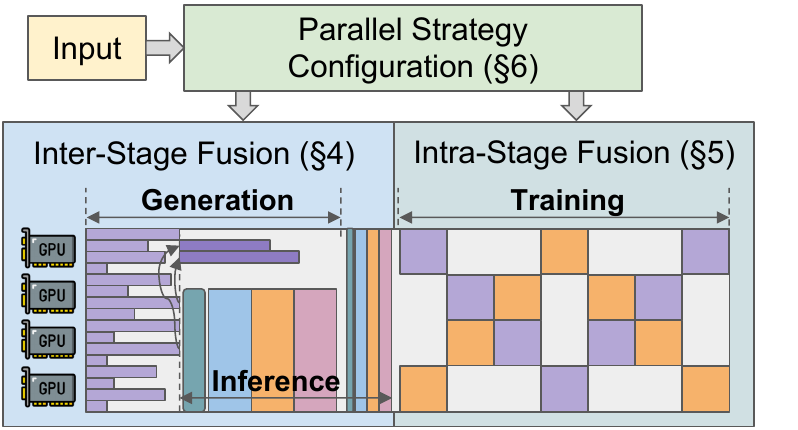}
    \vspace*{-0.05in}
    \caption{\sysname architecture.}
    \label{fig:overview:architecture}
\end{figure}

\vspace*{-0.2in}
\section{\sysname Overview}
\label{sec:overview}

To this end, we present \sysname to address
data skewness in the generation stage 
through data-aware inter-stage fusion (\S\ref{sec:inter_stage_fusing}) and 
mitigates pipeline bubbles in the 
training stage with model-aware intra-stage fusion (\S\ref{sec:intra_stage_fusing}).
Additionally, \sysname adopts a 
series of system optimizations (\S\ref{sec:system_opts})
tailored for RLHF training.
Here we provide a brief 
overview of \sysname as shown in Figure~\ref{fig:overview:architecture}.

\parabf{Workflow.}
With specific algorithm, model, and cluster configurations as input, \sysname first configures efficient 
parallel and deployment strategies following
the approach in ReaLHF~\cite{mei2024realhf} and HybridFlow~\cite{sheng2024hybridflow}. For each task within the RLHF workflow, 
\sysname assigns it a tailored 
parallel strategy to maximize GPU utilization.
At runtime, \sysname launches each task 
with its corresponding parallel strategy on the designated device mesh following the workflow dependency and handle the weight redistribution 
and data transmission between different tasks to ensure synchronous training semantics. 
Meanwhile, \sysname utilizes data-aware inter-stage fusion to fuse the generation and inference tasks and model-aware intra-stage fusion to fuse the training tasks.

\parabf{Inter-stage fusion.}
As depicted in the left part of Figure~\ref{fig:overview:architecture}, during the generation stage, there are multiple model instances, each 
managing complete model weights and part of the samples. \sysname actively monitors each generation instance and triggers 
migration when the number of remaining samples is below a particular threshold.
The remaining samples are 
migrated to some dedicated generation instances which are designated to handle 
long-tailed samples. Subsequently, the resources of the original generation 
instances are repurposed to launch inference tasks, 
thereby allowing for the overlap of inference tasks with the generation of 
long-tailed samples and fusing the execution of the two stages.

\parabf{Intra-stage fusion.}
As shown in the right part of Figure~\ref{fig:overview:architecture}, in the training stage, \sysname leverages the
insight that another inverse pipeline can complement current pipeline execution. It utilizes a lightweight algorithm to generate a fused pipeline 
schedule based on the model size and parallel strategies of the \texttt{Actor} and \texttt{Critic} model,
which minimizes the pipeline bubbles and activation memory usage.

\section{Data-aware Inter-Stage Fusion}
\label{sec:inter_stage_fusing}

In this section, we first analyze the opportunities of fusing the generation 
and inference stages (\S\ref{sec:inter_stage_fusing:opportunities}). 
We then introduce the fused execution plan to 
maximize the overlap between the two stages while preserving the data dependency
(\S\ref{sec:inter_stage_fusing:scheduling}).

\vspace{-0.1in}
\subsection{Opportunities and Challenges}
\label{sec:inter_stage_fusing:opportunities}

\vspace{-0.1in}
\parabf{Opportunities.}
The dependency between the two stages is predicated on the sample level,
implying that once a sample completes its generation stage, it can seamlessly advance to the inference stage.
This key observation motivates us to break the task into sample-level subtasks
without violating original synchronous training semantics.
Consequently, we can initiate inference tasks as soon as there exist completed samples in the generation stage.

\begin{figure}[t!]
    \centering
    \includegraphics[width=\linewidth]{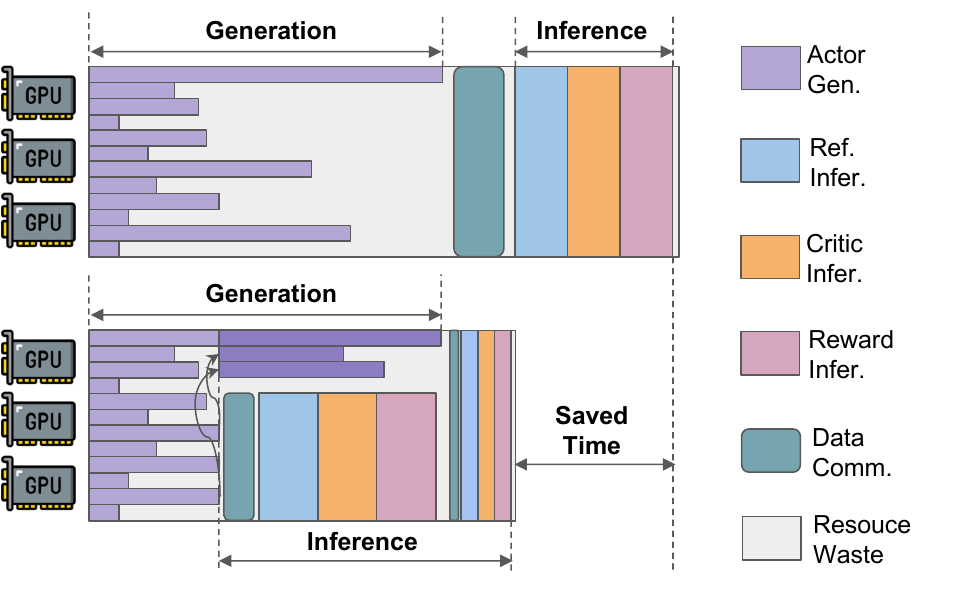}
    \vspace*{-0.25in}
    \caption{The timeline of serial (top) and fused (bottom) execution of generation and inference stages.}
    \label{fig:design:inter_stage_fusing}
\end{figure}

Based on this, we design a fused execution plan that overlaps the generation and inference stages.
Specifically, the fused plan first detects the point when the generation instance enters the long-tail decoding phase.
At this juncture, most of the samples have finished their generation stage but are waiting for the inference stage to start.
Each instance only leaves a few long-tailed samples to process, which is memory-bandwidth-bound as discussed in \S\ref{sec:motivation:problems}.
This condition facilitates the migration of samples from all instances to a few designated instances.
This migration not only consolidates the GPU utilization of the receiving instances
but also releases resources from the sending instances.
The released resources are then promptly utilized to launch inference tasks in advance.
One example of a fused execution plan is shown in Figure~\ref{fig:design:inter_stage_fusing}.
This approach greatly improves the GPU utilization and optimizes the overall execution time of the two stages.

\parabf{Challenges.}
With the inter-stage fusion, the problem now lies in generating an efficient
fused execution plan that decides the migration timing, destination, and mechanism.
These three factors significantly affect the overall execution time of the two stages,
and therefore must be chosen carefully.

Given that inter-stage fusion appears so promising, a natural question arises: 
can we break the synchronization between the inference and training stages similarly? Unfortunately, 
training tasks require each mini-batch to maintain the same data distribution, 
which necessitates random sampling from all the generated samples. To satisfy
this requirement, it implicitly introduces an unavoidable synchronization boundary.

\vspace{-0.1in}
\subsection{Fused Execution Plan}
\label{sec:inter_stage_fusing:scheduling}

The objective of the inter-stage fusion is to overlap the execution of the two stages as much as possible
without affecting the original generation task's execution time, thereby minimizing the overall execution time of the two stages.
Following this objective, we detail the migration triggering, destination, and mechanism below.

\parabf{Migration triggering.}
Triggering the migration significantly influences the efficiency of overlapping the two stages.
If migration is triggered too early, there remain excessive generation samples.
The migration results in only a few model instances remaining available to continue the generation task,
as most are redirected to handle inference tasks.
This scenario places an excessive load on the few generation instances,
leading to prolonged execution times for the original generation task and ultimately extending the overall process.
Conversely, if the migration is triggered too late, it loses most of the overlapping opportunity.
Based on the above analysis, we propose a migration threshold, denoted as $R_{t}$.
When \textit{the number of remaining samples in the generation stages falls below $R_{t}$}, the migration is triggered.
To optimize system performance and minimize total execution time, carefully determining $R_{t}$ is essential to
find the best trade-off.

To determine the optimal $R_{t}$, we initially conducted offline generation trials
on the training dataset to analyze
the length distribution of the generated samples. This analysis enables us to estimate the
computational load for both the generation and inference stages accurately and simulate the
execution time of the fused execution plan.
Due to the determinism and reproducibility of LLM computation, the simulation
technique for LLM training~\cite{bang2024vtrainsimulationframeworkevaluating,alpa} and inference~\cite{li2023alpaserve,zhong2024distserve} is well-studied.
We then systematically test $R_{t}$ values ranging from 5\% to 95\% of the global batch size,
simulating the overall execution time under each $R_{t}$. The optimal $R_{t}$ is identified as the value that
yields the minimum simulated execution time.
Additionally, during runtime, we refine the distribution by incorporating new generation samples,
which allows us to update $R_{t}$ as needed to continually optimize performance.

Note that it is possible to achieve finer-grained overlapping by triggering migration 
more than once. However, we find that in practice, the generation time of long-tailed samples
often exceeds the execution time of inference tasks, so one migration is usually sufficient
to fully overlap the entire inference stage, as shown in \S\ref{sec:eval:breakdown}.

\parabf{Migration destination.}
Once the migration timing is determined, we need to select $m$ generation
instances to handle the remaining samples. Assume there are $n$ generation instances in total,
then there will be $n - m$ instances to be repurposed for inference tasks.
To determine $m$, we revisit the computation characteristics of the generation task. 
During the decoding phase of generation, which is highly memory-bandwidth-bound, 
as long as the batch size does not exceed a certain threshold that saturates the GPU 
(denoted as $BS_{max}$), the latency of each decoding iteration remains almost constant~\cite{zhong2024distserve,wu2023fast}
thanks to the massive parallel computing units of modern GPUs.
The value of $BS_{max}$ depends on the specific GPU hardware and can be determined through prior profiling.
Based on this, we set the first constraint as  
$m \geq \frac{R_{t}}{BS_{max}}$,
which ensures that the generation time of long-tailed
samples remain unchanged as before the migration.

Another factor to consider is the memory constraint.
During LLM generation, the key-value cache is maintained for each token position for
autoregressive generation.
When the sequence length is long, this memory consumption becomes non-negligible~\cite{kwon2023efficient}.
Therefore, we need to ensure that the target instances have sufficient memory to 
accommodate the remaining long-tailed samples, avoiding out-of-memory 
issue or blocking the sample processing. Thus, we set the second constraint
as $m \geq \frac{R_{t} * M}{C}$, where $M$ denotes the key-value cache consumption of the sample with the maximum
output length and $C$ is the available GPU memory of the target instance allocated for key-value cache.
In summary, the final $m$ is determined by the maximum of the two constraints.

Next, we need to determine which $m$ out of the
$n$ generation instances to process the remaining
long-tailed samples. To minimize migration overhead,
we select the top $m$ instances that have the most samples remaining.
This strategy minimizes the total number of samples requiring migration.

\parabf{Migration mechanism.} To migrate an unfinished sample to the target instance,
we have two choices. The first approach is to transfer the generated key-value 
cache to the target instance over the network, allowing it to continue the subsequent 
generation immediately upon receipt. The overhead in this case primarily comes from the 
network transmission. The second approach is to discard the key-value cache and 
only transmit the generated tokens of the sample. 
Compared to the first approach, this method incurs minimal network latency but 
comes at the cost of rerunning the prefill phase to generate the key-value cache.
The choice of migration mechanism depends on the GPU hardware and network bandwidth.
In our experiments, thanks to the high-bandwidth RDMA connections,
we choose the first approach and the migration overhead is 
negligible compared to the iteration time, as shown in \S\ref{sec:eval:breakdown}.

\begin{figure*}[t!]
    \centering
    \includegraphics[width=0.95\linewidth]{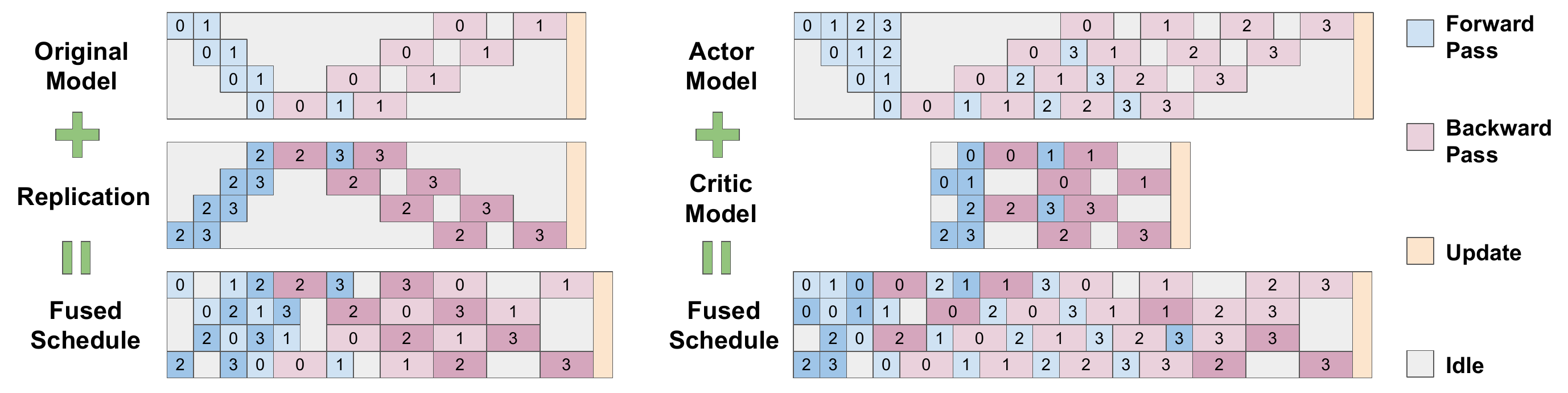}
    \hspace*{0.25in}(a) Chimera~\cite{li2021chimera} \hspace*{2.2in} (b) \sysname \hspace*{0.9in}
    \vspace*{-0.1in}
    \caption{(a) An example of the symmetric bi-directional pipeline schedule used in Chimera~\cite{li2021chimera} with four pipeline stages and four micro-batchs.
    (b) An example to show our method can fuse two different models with different numbers of pipeline stages.}
    \vspace*{-0.2in}
    \label{fig:design:intra_stage_fusing}
\end{figure*}

After completing the migration, we can release the GPU resources 
of the remaining $n - m$ generation instances to launch inference tasks.
This process involves weights redistribution and data transmission to prepare
the context of inference tasks, which we discuss in \S\ref{sec:system_opts}. 
Once the inference tasks are launched, the remaining long-tailed 
samples can be streamed to the inference instances as soon as their generation stage
is completed, allowing for seamless processing.
Additionally, if the long-tail generation task finishes first, 
we release its resources for the inference tasks.

\vspace{-0.1in}
\section{Model-aware Intra-Stage Fusion}
\label{sec:intra_stage_fusing}

In this section, we first analyze the opportunities of fusing the
two tasks in the training stage (\S\ref{sec:intra_stage_fusing:opportunities}).
Then we propose a lightweight algorithm to generate the fused pipeline schedule to minimize
the pipeline bubbles and memory usage (\S\ref{sec:intra_stage_fusing:schedule}).

\vspace{-0.1in}
\subsection{Opportunities and Challenges}
\label{sec:intra_stage_fusing:opportunities}

\paraf{Opportunities.}
During the RLHF training stage, the \texttt{Actor} and \texttt{Critic} models are trained
independently. Inspired by the bi-directional pipeline schedule~\cite{li2021chimera} from Chimera,
we can break the two training tasks into subtasks of micro-batches and co-locate
these subtasks to mutually fill the pipeline bubbles of each other.
Originally, Chimera enhances single-model training by replicating the model and trains
the replicated model with 1F1B schedule in opposite pipeline directions, as depicted in Figure~\ref{fig:design:intra_stage_fusing}(a).
This approach not only mitigates pipeline bubbles but also brings a balanced distribution of activation memory.
RLHF training stage inherently involves two distinct models,
enabling us to apply the fused pipeline schedule without extra model replication.

\parabf{Challenges.}
In Chimera, the bi-directional pipeline schedule utilizes one identical model replica
with uniform size and parallel strategy, resulting in a symmetric execution flow.
In contrast, RLHF involves training heterogeneous \texttt{Actor} and \texttt{Critic} models,
which differ not only in size but also in their optimal parallel strategies.
Consequently, the symmetric schedule from Chimera is no longer applicable in 
the face of model heterogeneity in terms of size and parallel configuration.
In the following, we introduce a lightweight and efficient algorithm to automatically generate 
the fused pipeline schedule under this more generalized setting.

\vspace{-0.1in}
\subsection{Fused Pipeline Schedule}
\label{sec:intra_stage_fusing:schedule}

\begin{table}[h]
    \centering
    \begin{tabular}{|l||l|}
        \hline
        \textbf{Symbol} & \textbf{Description} \\ \hline
        $S$ & The fused pipeline schedule. \\ \hline
        $l_{ij}$ & The latency of subtask $S_{ij}$. \\ \hline
        $C$ & The activation memory capacity of each stage. \\ \hline
        $K_1$ & The fusion factor of model $A$. \\ \hline
        $K_2$ & The fusion factor of model $B$. \\ \hline
        $N$ & The total number of pipeline stages to be fused. \\ \hline
        $N_1$ & The number of pipeline stages of model $A$. \\ \hline
        $N_2$ & The number of pipeline stages of model $B$. \\ \hline
        $M_1$ & The number of micro-batches for model $A$. \\ \hline
        $M_2$ & The number of micro-batches for model $B$. \\ \hline
    \end{tabular}
    \vspace*{-0.1in}
    \caption{Key notations in problem formulation.}
    \label{design:tab:notations}
\end{table}

\paraf{Problem transformation.}
Assume the parallel strategies for models $A$ and $B$ are $(dp_1, pp_1, tp_1)$
and $(dp_2, pp_2, tp_2)$, which denotes the parallel degree of each 3D-parallelism dimension. Each will utilize all the GPUs in the cluster.
We require that $tp$ is the powers of two, which is 
commonly adopted in practice. 
For the case where $tp_1 \neq tp_2$, the challenge for a fused pipeline schedule
lies in the fact that the pipeline stage of the two models contains a different number of GPUs.
Without loss of generality, we assume $tp_1 = s \times tp_2$.
In this situation, we merge every $s$ consecutive pipeline stages of model $B$
into one stage, redividing model $B$ into $\frac{pp_2}{s}$ pipeline stages.
This ensures that each stage of both models uses the same number of GPU resources. Note
that here we require $pp_2$ to be divisible by $s$, which is easy to realize in practice.
After that, we can transform the problem into fusing
$K_1$ pipeline groups of model $A$ with $K_2$ pipeline groups
of model $B$, where $K_i$ denotes the fusion factor and $K_1$ and $K_2$ are coprime.
Figure~\ref{fig:design:intra_stage_fusing}(b) illustrates an example of $(K_1, K_2) = (1, 2)$.
Next, we precisely define the fused pipeline schedule and formulate
the problem. Table~\ref{design:tab:notations} summarizes the key notations.

\parabf{Problem formulation.}
We assume the two models to be fused
each has $N_1$ and $N_2$ pipeline stages ($K_1 \times N_1 = K2 \times N_2 = N$),
and each pipeline needs to process $M_1$ and $M_2$ micro-batches, respectively. Since the global batch size is fixed,
we have $K_1 \times M_1 = K2 \times M_2$.
The fused pipeline schedule is represented as a matrix $S$,
where $S_{ij}$ represents the $j$-th subtask (micro-batch) to be scheduled in the $i$-th stage.
Since there are $N$ stages in total and each stage needs to process the forward and backward computation for all 
micro-batches of the two models exactly once, the shape of $S$ is $N \times 2(M_1 + M_2)$.
The latency of subtask $S_{ij}$ can be profiled in advance and is denoted as $l_{ij}$.
Note that not all $S$ can be scheduled. A valid schedule must satisfy the 
following constraints:
\begin{itemize}[leftmargin=*]
    \item[1.] Data dependency: the forward and backward of the same 
    micro-batch across different stages must be executed following the original data dependency.
    \item[2.] Deadlock avoidance: there exists no cycles in the overall dependency graph to avoid deadlocks.
    \item[3.] Memory constraint: the peak activation memory usage of each stage must be less than 
    $C$ to avoid out-of-memory.
\end{itemize}
The main optimization target is to find a valid schedule $S$ that has the minimum latency.
Additionally, among schedules with the same latency, we prefer the ones with
lower peak activation memory to optimize resource usage.

\begin{algorithm}[t]
    \caption{Generate Fused Pipeline Schedule.}
    \label{alg:design:main_routine}
    \begin{algorithmic}[1]
        \Function {GenerateFusedPipelineSchedule}{$S_0$}
            \State $s_{current} \gets S_0$
            \State $e_{current} \gets \mathit{ComputeEnergy}(s)$
            \State $T \gets e_{current}$
            \State $s^* \gets s_{current}$
            \State $e^* \gets e_{current}$
            \While{$T > \epsilon$}
                \State $s_{neighbor} \gets \mathit{ComputeNeighbor(s_{current})}$
                \State $e_{neighbor} \gets \mathit{ComputeEnergy(s_{neighbor})}$
                \If{$e_{neighbor} < e^*$}
                    \State $s^* \gets s_{neighbor}$
                    \State $e^* \gets e_{neighbor}$
                \EndIf
                \If{$P(e_{current}, e_{neighbor}, T) > Rand(0, 1)$}
                    \State $s_{current} \gets s_{neighbor}$
                    \State $e_{current} \gets e_{neighbor}$
                \EndIf
            \State $T \gets T \times \alpha$
            \EndWhile
            \State \Return $s^*$
        \EndFunction
    \end{algorithmic}
\end{algorithm}

\parabf{Algorithm overview:} The problem has many constraints and variables.
A naive solution is to extend the bi-directional pipeline~\cite{li2021chimera} greedily which always schedules feasible micro-batches.
If the micro-batches of two models are both ready, it favors the larger model,
with the expectation that the smaller one can flexibly fill in the bubbles later.
However, the greedy method lacks a global perspective. For example, 
certain micro-batches could be deliberately delayed to fill the bubbles 
later. Moreover, it provides no optimization for memory usage.
Thus, the greedy approach does not yield optimal performance,
as we will demonstrate in \S\ref{sec:eval:ablation}.

Instead, we use simulated annealing~\cite{kirkpatrick1983optimization} to search for a better solution.
The motivation for using simulated annealing is that we have a huge search space
with many variables. Simulated annealing is effective in finding acceptable
local optimums in a reasonable amount of time while finding the global optimum 
is computationally expensive.
Furthermore, we will show that our method can achieve
the theoretical lower bound most of the time as shown in \S\ref{sec:eval:ablation}.

At a high level, we use the fused pipeline schedule $S$ as the state in simulated
annealing. We use the schedule found by the greedy algorithm mentioned above as the
initial state and probabilistically jump to a neighbor state in each iteration,
aiming to find a schedule with the lowest latency.
Our approach has two benefits. First, it is easy to scale out. We can perform searching using different random seeds across 
hundreds of CPU cores and select the best result among them. Second,
it can easily extend to more models, which arises in multimodal~\cite{zhang2024disttrainaddressingmodeldata}
and multi-agent~\cite{sun2024llm} training scenarios.
Now we describe the algorithm in more detail below.

\begin{algorithm}[t]
    \caption{Generate A Random Neighbor Schedule.}
    \label{alg:design:neighbor_routine}
    \begin{algorithmic}[1]
        \Function {ComputeNeighbor}{$S$}
            \While{$True$}
                \State $i \gets RandInt(1, N)$
                \State $j \gets RandInt(1, 2(M_1 + M_2)-1)$
                \State $S' = Swap(S_{ij}, S_{i(j+1)})$
                \If{$CheckValid(S')$}
                    \State \Return $S'$
                \EndIf
            \EndWhile
        \EndFunction
    \end{algorithmic}
\end{algorithm}

\parabf{Simulated Annealing (Algorithm~\ref{alg:design:main_routine}):} The algorithm uses 
the solution found by the greedy algorithm as the initial state $S_0$, and its execution time as the initial temperature (line 2-3). $s^*$
is used to store the schedule with the lowest latency and $e^*$ is the energy (latency) of $s^*$.
The algorithm searches until temperature $T$ is less than an epsilon value (lines 7-16). $T$ is decreased
by a factor of $\alpha$ in every iteration. At each iteration, it uses $ComputeNeighbor$
subroutine to find a neighbor state of the current one and uses $ComputeEnergy$ to compute the
energy of the neighbor state. If the neighbor state has lower energy than $s^*$,
it updates $s^*$ (lines 10-12). The algorithm uses a probabilistic function $P$ to decide
whether to transition from the current state to the neighbor state.
The probabilistic function $P$ is defined as
follows: if the neighbor state has a lower energy than the current state, the probability is
$1$; otherwise, the probability is $e^{(e_{current}-e_{neighbor})/T}$. 

\begin{algorithm}[t]
    \caption{Compute Energy}
    \label{alg:design:energy_routine}
    \begin{algorithmic}[1]
        \Function {ComputeEnergy}{$S$}
            \State \Return $max_{1\leq i\leq N}$ComputeFinishTime($S_{i,M}$)
        \EndFunction

        \Function {ComputeFinishTime}{$S_{ij}$}
            \If{$S_{ij} \in Memo$}
                \Return $Memo.get(S_{ij})$
            \EndIf
            \State $S^{intra} \gets GetIntraDependency(S_{ij})$
            \State $S^{inter} \gets GetInterDependency(S_{ij})$
            \State $IntraTime \gets ComputeFinishTime(S^{intra})$
            \State $InterTime \gets ComputeFinishTime(S^{inter})$
            \State $FinishTime = max(IntraTime, InterTime) + l_{ij}$
            \State $Memo.put(S_{ij}, FinishTime)$
            \State \Return $FinishTime$
        \EndFunction
    \end{algorithmic}
\end{algorithm}

\parabf{ComputeNeighbor (Algorithm~\ref{alg:design:neighbor_routine}):} This routine finds
a neighbor state of the current state. It randomly swaps two adjacent subtasks from a random stage.
If the neighbor state is invalid, it will undo the change and repeat the random swapping until finding a valid schedule.
We select adjacent subtasks rather than any two subtasks in one stage primarily to control the degree 
of disturbance, which is found to be more effective in practice.

\parabf{ComputeEnergy (Algorithm~\ref{alg:design:energy_routine}):}
This function computes the execution time for a given valid schedule $S$.
We use a memoized recursion to calculate the finish time of the last subtask
in each stage and pick the maximum as the execution time. For subtask $S_{ij}$, its start time is decided by two
dependencies: inter-stage data dependency and intra-stage data dependency.
The inter-stage dependency is the completion of subtask in the upstream pipeline stage
corresponding to the same micro-batch.
The intra-stage dependency is the completion of the preceding subtask in the same pipeline stage.
Consequently, the end time of $S_{ij}$ is
determined by the maximum of these two dependencies plus its own computation time, $l_{ij}$.
The detailed algorithm is shown in Algorithm~\ref{alg:design:energy_routine}.
It has time complexity of $O(N \times (M_1 + M_2))$.

\parabf{Optimizing memory usage.} When we obtain a latency-optimized solution
$S^*$, we use it as the initial state to run another round of simulated annealing
similar to Algorithm~\ref{alg:design:main_routine}. However, this time
we replace the $ComputeEnergy$ function with one that calculates the
peak activation memory for the given schedule. 
Additionally, we only allow state transitions when the neighbor's latency
does not degrade. In this way, we can achieve a solution that not only 
has promising latency but also optimizes the activation memory usage.

\vspace{-0.1in}
\section{Implementation}
\label{sec:system_opts}

We implement \sysname based on Megatron-LM~\cite{shoeybi2020megatronlm} with 7K lines of code
in Python, C++, and CUDA. Megatron-LM applies 3D-parallelism
for single-model training. We extend it to support multiple device meshes to launch 
different tasks asynchronously with tailored parallelism and deployment strategy.

\parabf{Parallel strategy configuration.}
Optimizing the parallel strategy for LLM training task is a well-studied problem~\cite{alpa,shoeybi2020megatronlm,hetu,mei2024realhf}.
Due to the deterministic nature of LLM computation~\cite{gujarati2020serving}, the execution time and memory cost can be accurately
modeled through minimal profiling~\cite{bang2024vtrainsimulationframeworkevaluating,li2023alpaserve,zhong2024distserve}. An optimal solution can then be 
found through optimizations~\cite{alpa,pipedream,hetu,mei2024realhf}. 
The generation and inference task involves only the forward pass, 
which can be modeled as a simplified problem of LLM training~\cite{li2023alpaserve,zhong2024distserve}.
ReaLHF~\cite{mei2024realhf} is a state-of-the-art solution
that optimizes the parallel strategy for RLHF tasks and we adopt a similar model-then-optimize approach.
We build a simulator to accurately estimate the runtime statistics of the RLHF task
under a specific parallel strategy and workload pattern. Then we follow the
guidelines in the Megatron-LM~\cite{shoeybi2020megatronlm} paper to prune the design space and brute-force search 
the optimal strategy for each RLHF task.

\parabf{Inter-stage fusion.}
We extend the simulator mentioned above from the task level to
the workflow level. Given the parallel strategies, workload pattern, and migration threshold
$R_t$, it considers the task dependencies and data transmission overhead
and simulates the overall execution time of the generation and inference stages.
Then we can choose the best migration threshold $R_t$ to minimize the fused execution time.

\parabf{Intra-stage fusion.}
We use MPI~\cite{gabriel2004open}
to parallelize the simulated annealing algorithm used for fused pipeline schedule generation (\S\ref{sec:intra_stage_fusing:schedule}).
Under different random seeds, the computations are completely independent, so that
we can easily scale the computation to hundreds of CPU cores, making it highly likely to find the optimal solution (\S\ref{sec:eval:ablation}).
After finding the best solution, it will generate the 
sequences of NCCL~\cite{nccl} operations on each device following the fused pipeline schedule.
The underlying execution engine will follow the instruction flow to execute the exact schedule during runtime. The code of the intra-stage fusion algorithm is publicly available\footnote{https://github.com/FlexFusion/FlexFusion}.

\parabf{System optimizations.}
We adopt a series of system optimizations tailored for RLHF training. 
Since the RLHF training workflow is well-known, many of the following optimizations
are also implemented or partially supported in a similar way in other
RLHF frameworks like ReaLHF~\cite{mei2024realhf}, HybridFlow~\cite{sheng2024hybridflow}, OpenRLHF~\cite{hu2024openrlhf},
DeepSpeed-Chat~\cite{yao2023deepspeedchateasyfastaffordable}, and PUZZLE~\cite{lei2024puzzle}.

For the generation stage, we implement an in-house inference engine which integrates most of
the modern techniques optimized for LLM autoregressive generation like continuous batching~\cite{yu2022orca},
prefix sharing~\cite{zheng2024sglangefficientexecutionstructured}, and chunked-prefill~\cite{sarathi}.
For the inference stage, we optimize the computation for
Generalized Advantage Estimation~\cite{schulman2018highdimensionalcontinuouscontrolusing},
which unrolls its original recursive formula along the output length dimension
and transforms the recursive computation into a single matrix multiplication,
greatly reducing the kernel launch overhead.
For the training stage, we evenly distribute each mini-batch across the DP groups~\cite{mei2024realhf}
based on the sequence length of the samples to ensure that the workloads are
roughly balanced across the DP groups, effectively tackling the straggler issue.
To reduce the task switching overhead~\cite{lei2024puzzle,sheng2024hybridflow,mei2024realhf}, we minimize the cross-node communication~\cite{zhuang2024optimizingcommunicationmodelparallelism} for
the \texttt{Actor} and \texttt{Critic} models whose latest weights after the training stage
need to be redistributed to the appropriate devices according to the new parallel strategies. 
For the \texttt{Ref} and \texttt{RW} models whose weights remain unchanged,
we keep them in CPU memory and swap them into GPU memory as needed~\cite{sheng2024hybridflow,mei2024realhf},
overlapping with the computation of previous tasks.

To separate the differences in low-level system implementations from the core
techniques (inter-stage fusion in~\S\ref{sec:inter_stage_fusing} and intra-stage
fusion in~\S\ref{sec:intra_stage_fusing}), our evaluation in~\S\ref{sec:evaluation}
uses \sysname-Base that includes these optimizations but
without inter- and intra-stage fusion as an additional baseline.

\begin{figure*}[t!]
    \centering
    \includegraphics[width=0.95\linewidth]{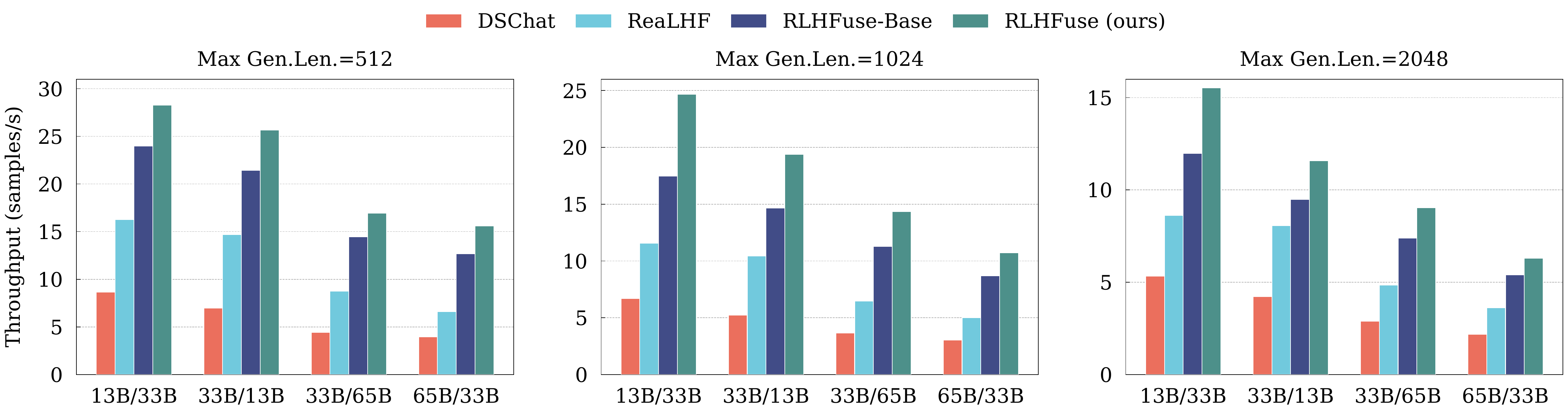}
    \vspace*{-0.1in}
    \caption{End-to-end throughput of RLHF training systems under different
    generation and model size settings.}
    \vspace*{-0.2in}
    \label{fig:eval:end2end}
\end{figure*}

\vspace{-0.1in}
\section{Evaluation}
\label{sec:evaluation}

In this section, we evaluate \sysname under different sizes of
LLMs ranging from 13B to 65B on real-world datasets.

\parabf{Cluster testbed.} We deploy \sysname on a production cluster for RLHF training 
with 32 nodes and 256 GPUs. Each node has 2TB of memory and 8 NVIDIA Hopper GPUs connected with NVLINK. 
Nodes are connected by $8*200$ Gbps RDMA network based on RoCEv2 with rail-optimized topology. 
The simulated annealing algorithm in~\S\ref{sec:intra_stage_fusing} is conducted on four
CPU nodes, each with 2 AMD 9654 CPUs and there are 768 physical cores in total.

\parabf{Models and datasets.} Following previous work~\cite{lei2024puzzle},
we choose the LLaMA models~\cite{touvron2023llama1} ranging from 13B to 65B,
which is a popular LLM family used in academia and industry. The
detailed specifications are listed in Table~\ref{eval:tab:configs}.
We use the HH-RLHF~\cite{bai2022traininghelpfulharmlessassistant} dataset,
which is open-sourced by Anthropic to train a helpful and harmless assistant 
with RLHF. 

\parabf{Settings.} Since the \texttt{Actor} and \texttt{Ref} models
are of the same size, as are the \texttt{Critic} and \texttt{RW} models,
we choose four different model size settings for \texttt{Actor/Critic} pair
: 13B/33B, 33B/13B, 33B/65B, and 65B/33B.
We also vary the maximum output length to see the performance
under different generation settings.
In each training iteration, we use a global batch size of 512,
a mini-batch size of 64, and take one gradient step per mini-batch following the
LLaMA technical report~\cite{touvron2023llama2}.

\parabf{Metrics.} For the end-to-end experiment, we measure the \textit{sample throughput}
following previous work~\cite{lei2024puzzle}. Sample throughput is defined as the average number
of samples processed per second. Under each setting, we record the sample throughput
over 20 consecutive training iterations after warm-up.

\subsection{End-to-End Results}
\label{sec:eval:end_to_end}

\begin{table}[t!]
    \centering
    \resizebox{0.8\linewidth}{!} {
    \begin{tabular}{cccccc}
        \toprule
        \multirow{2}{*}{\textbf{Models}} & \textbf{\# of} & \textbf{\# of} & \textbf{Hidden} &
        \textbf{Intermediate} \\
         & \textbf{Layers} & \textbf{Heads} & \textbf{Size} & \textbf{Size} \\
        \midrule
        LLaMA-13B & 40 & 40 & 5120 & 20480 \\
        LLaMA-33B & 60 & 52 & 6656 & 26624 \\
        LLaMA-65B & 80 & 64 & 8192 & 32768 \\
        \bottomrule
    \end{tabular}
    }
    \vspace{-0.1in}
    \caption{LLM specifications.}
    \label{eval:tab:configs}
\end{table}

We compare the end-to-end performance of \sysname against
the following RLHF training frameworks.

\begin{itemize}[leftmargin=*]
    \item \textbf{DeepSpeed-Chat~\cite{yao2023deepspeedchateasyfastaffordable}} (DSChat)
    colocates all the models on the same set of devices and only supports ZeRO-3 data parallelism~\cite{zero}
    during training. It utilizes a \textit{HybridEngine} to switch from 
    ZeRO-3 DP to TP in the generation stage.
    Since each GPU requires at least one sample during training and it only supports ZeRO-3 DP,
    we increase its mini-batch size to 256
    while maintaining the original global batch size in order to run it successfully on our testbed.
    Note that with a larger mini-batch size, this adjustment is more favorable for its throughput performance.
    \item \textbf{ReaLHF~\cite{mei2024realhf}} proposes parameter reallocation to
    flexibly redistribute parameters between tasks, enabling tailored
    3D-parallel strategy for each task.
    However, as discussed in \S\ref{sec:motivation:problems},
    it suffers from the issues of data skewness and pipeline bubbles
    without employing the subtask-level optimizations.
    \item \textbf{\sysname-Base} is \sysname without inter- and intra-stage
    fusion but with all the system optimizations enabled as described in \S\ref{sec:system_opts}.
    We include this baseline to demonstrate the performance improvements
    brought by our fusion techniques in \S\ref{sec:inter_stage_fusing} and \S\ref{sec:intra_stage_fusing},
    eliminating any unfair comparisons caused by the differences in underlying framework implementations
    and other optimization techniques in LLM generation and training which are orthogonal to the proposed stage fusion techniques.
\end{itemize}
\begin{figure*}[t!]
    \centering
    \includegraphics[width=0.98\linewidth]{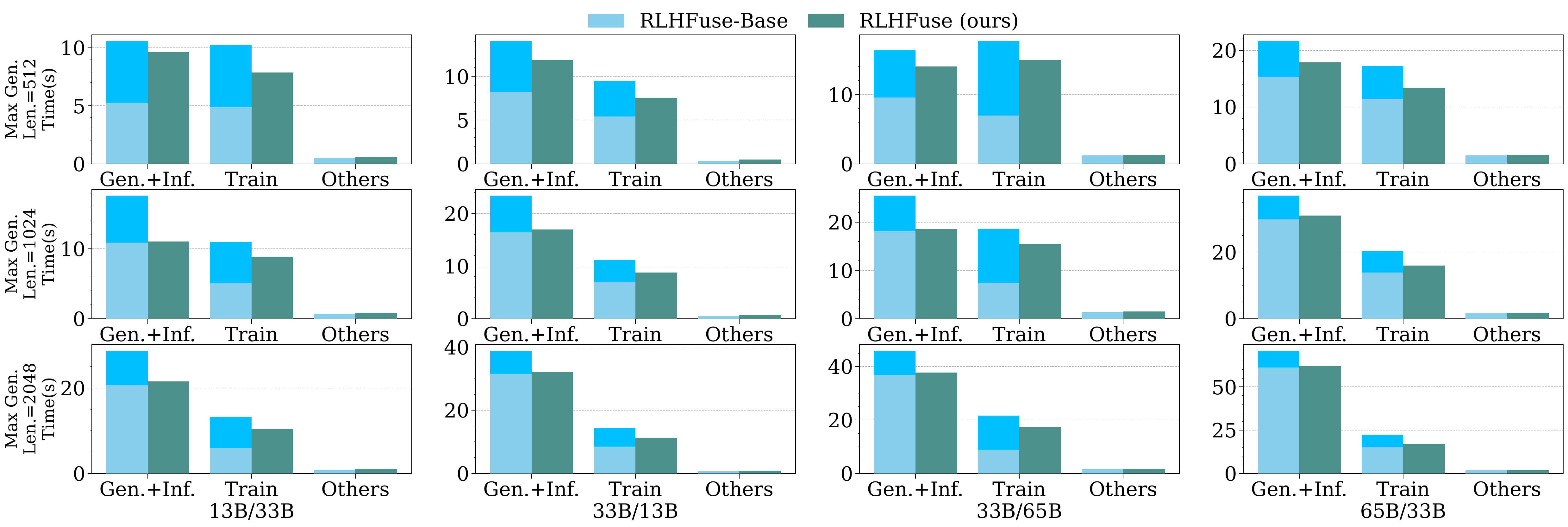}
    \vspace*{-0.1in}
    \caption{The RLHF iteration breakdown under different generation and model size settings.}
    \vspace*{-0.2in}
    \label{fig:eval:breakdown}
\end{figure*}
Figure~\ref{fig:eval:end2end} shows the end-to-end performance of the RLHF training
systems under three generation settings. We set the maximum generation length to 512, 1024, and 2048, respectively.
Under a specific generation setting, the configuration X/Y denotes parameter counts in actor and critic models,
mirrored in the reference and reward models, respectively. 
Compared to DSChat, \sysname achieves $2.5\times$--$3.7\times$ higher throughput.
This is because DeepSpeed-Chat colocates all the models on the same set of 
GPUs, making it impossible to adopt the most suitable parallel strategy for each 
task, which introduces additional computation and communication overhead.
ReaLHF implements a flexible execution plan generator, which selects the most efficient
3D-parallel strategy for each task, such as balancing between DP and PP sizes to reduce the number of 
pipeline stages while avoiding OOM, thereby partially mitigating pipeline bubbles.
However, \sysname goes further by eliminating the vast majority of bubbles and addressing
the long-tail issue in the generation stage with intra- and inter-stage fusion. Additionally, \sysname integrates
production-grade optimizations in the underlying framework
tailored for large-scale training scenarios (\S\ref{sec:system_opts}).
As a result, \sysname achieves $1.4\times$--$2.4\times$ higher throughput compared to ReaLHF.
Compared to \sysname-Base, our system achieves a relative improvement
of $1.2\times$--$1.4\times$ on the throughput.
This improvement is entirely from stage fusion,
which significantly alleviates the issues of data skewness in the generation stage
and pipeline bubbles in the training stage.
In the following, we conduct a detailed breakdown analysis
to show the performance improvement.

\vspace{-0.2in}
\subsection{Performance Analysis}
\label{sec:eval:breakdown}

To further understand the performance improvement of \sysname, we show its
RLHF iteration breakdown and compare it with \sysname-Base.
As shown in Figure~\ref{fig:eval:breakdown}, each row represents a generation
setting, and each column represents a model setting X/Y. We divide one RLHF
iteration into three main parts: the generation plus inference stage
(\texttt{Gen.+Inf.}), the training stage (\texttt{Train}),
and all other overheads (\texttt{Others}), such as the data transmission and weights redistribution time.

For \texttt{Gen.+Inf.}, since \sysname-Base does not perform inter-stage fusion
and executes the two stages in serial,
we use light and dark colors to represent the time spent on the generation
and inference stages respectively to better show the speedup of inter-stage fusion.
It can be seen that as the maximum generation length increases, the processing time
of long-tailed samples is long enough for \sysname to fully overlap
the inference stage execution, achieving $1.2\times$--$1.6\times$ speedup.
For the training stage, similarly, we use light and dark colors to represent the
\texttt{Actor} and \texttt{Critic} training time in \sysname-Base.
Through intra-stage fusion, \sysname greatly mitigates pipeline bubbles
and reduces the execution time of the training stage by $1.2\times$--$1.3\times$.
As for the other overhead, thanks to our optimizations in task switching (\S\ref{sec:system_opts}),
it only accounts for less than 3\% of the total iteration time,
and the migration overhead of inter-stage fusion is also negligible.

\begin{figure}[t!]
    \centering
    \includegraphics[width=0.95\linewidth]{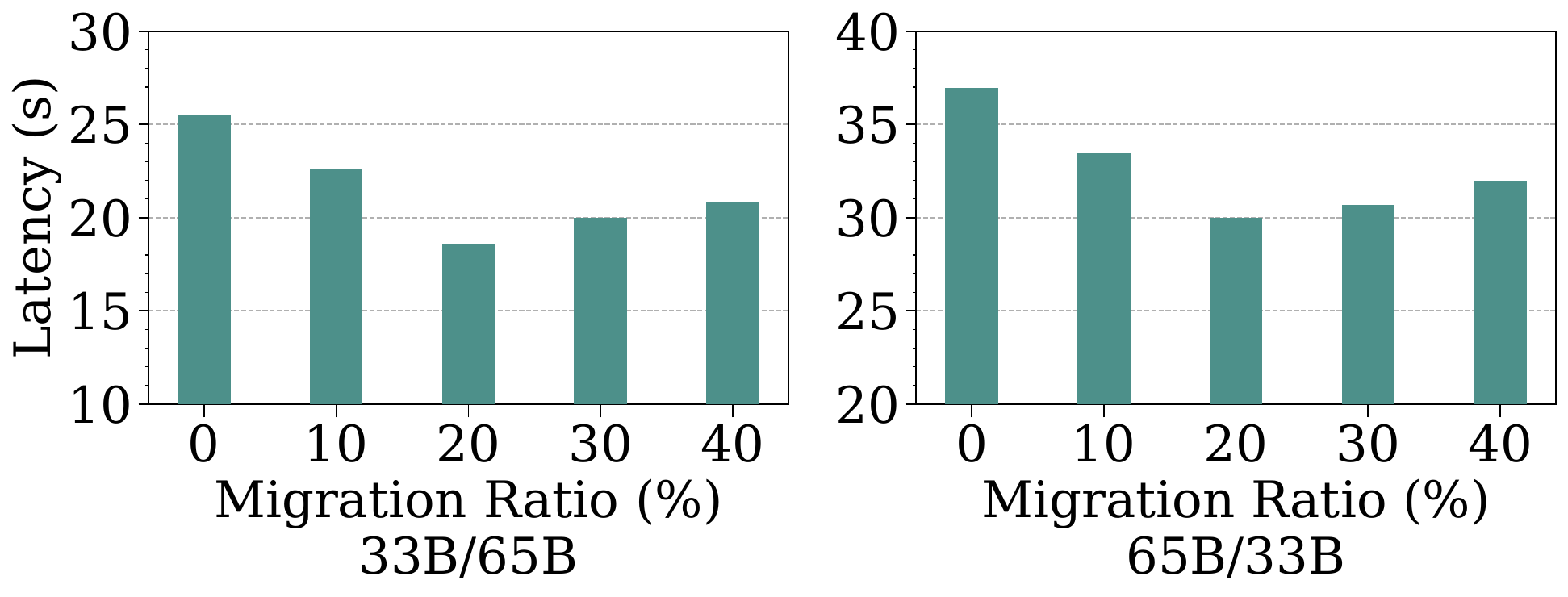}
    \caption{The fused generation and inference stage execution time of \sysname under different
    migration ratios and model settings. The maximum generation length is set to 1024.}
    \label{fig:eval:ablation}
\end{figure}

\subsection{Effectiveness of Stage Fusion}
\label{sec:eval:ablation}
In this section, we show the effectiveness of
\sysname's key techniques in \S\ref{sec:inter_stage_fusing} and \S\ref{sec:intra_stage_fusing}.

\parabf{Inter-stage fusion.} One key parameter in inter-stage fusion is the
migration threshold $R_t$. As $R_t$ increases from zero, it shifts from
fully serial execution to more aggressive fused execution. However, if $R_t$ is set too
large, the long-tail generation instances may be overloaded,
extending the original generation time. We measure the execution time of the fused generation and inference stage
under different migration ratios ($R_t / BS$) as shown in Figure~\ref{fig:eval:ablation}.
We can see the optimal latency is achieved when the remaining samples account for about 20\% of the batch size.
Note that $R_t$ is related to the output length distribution of the model
and this experiment is conducted during the early stage of training.
As training progresses, $R_t$ requires periodic adjustments to
adapt to the distribution change. In practice, such changes are usually not
too drastic, so the adjustment frequency remains low.

\begin{table}[t!]
    \centering
    \resizebox{\linewidth}{!} {
    \Huge
    \begin{tabular}{c|c|c|c|c|c|c|c|c|c}
        \toprule
        \multicolumn{4}{c|}{\multirow{2}{*}{\textbf{Settings}}} & \multicolumn{4}{c|}{\textbf{Latency Speedup}} & \multicolumn{2}{c}{\textbf{Peak Memory}} \\
                                          \multicolumn{4}{c|}{} & \multicolumn{4}{c|}{\textbf{relative to 1F1B}} & \multicolumn{2}{c}{\textbf{relative to 1F1B}} \\
        \midrule
        \textbf{Models} & \textbf{PP0} & \textbf{PP1} & \textbf{GBS} & \textbf{1F1B+} & \textbf{Greedy} & \textbf{Ours} & \textbf{LB} & \textbf{Greedy} & \textbf{Ours} \\
        \cline{1-10}
        \multirow{6}{*}{33B/13B} & \multirow{6}{*}{8} & \multirow{3}{*}{4}   & 8  & 1.10 & 1.29 & \textbf{1.38} & \textbf{1.38} & 1.51 & \textbf{1.0}      \\
                                 &                    &                      & 16 & 1.06 & 1.12 & \textbf{1.30} & \textbf{1.30} & 2.14 & \textbf{1.0}      \\
                                 &                    &                      & 32 & 1.03 & 1.10 & \textbf{1.15} & \textbf{1.15} & 2.75 & 1.26   \\
                                 \cline{3-10}
                                 &                    & \multirow{3}{*}{8}   & 8  & 1.07 & \textbf{1.4}  & \textbf{1.4}  & \textbf{1.4}  & 1.51 & \textbf{1.0}      \\
                                 &                    &                      & 16 & 1.05 & 1.15 & \textbf{1.32} & \textbf{1.32} & 2.00 & 1.19   \\
                                 &                    &                      & 32 & 1.03 & 1.08 & \textbf{1.17} & \textbf{1.17} & 2.75 & 1.31   \\ \hline
        \multirow{6}{*}{65B/33B} & \multirow{6}{*}{16} & \multirow{3}{*}{8}  & 16 & 1.05 & 1.28 & \textbf{1.48} & \textbf{1.48} & 1.61 & \textbf{1.0}      \\
                                 &                    &                      & 32 & 1.03 & 1.14 & \textbf{1.27} & \textbf{1.27} & 2.23 & \textbf{1.0}      \\
                                 &                    &                      & 64 & 1.02 & 1.12 & \textbf{1.15} & \textbf{1.15} & 2.88 & 1.26   \\
                                 \cline{3-10}
                                 &                    & \multirow{3}{*}{16}  & 16 & 1.03 & 1.27 & \textbf{1.5}  & \textbf{1.5}  & 1.57 & \textbf{1.0}      \\
                                 &                    &                      & 32 & 1.02 & 1.16 & \textbf{1.33} & \textbf{1.33} & 2.00 & 1.22   \\
                                 &                    &                      & 64 & 1.01 & 1.09 & 1.16 & \textbf{1.17} & 2.88 & 1.47   \\
        \bottomrule
    \end{tabular}
    }
    \vspace{-0.05in}
    \caption{The latency speedup and peak activation memory cost of pipeline schedules found by different algorithms
    under different models, pipeline stages, and global batch size settings.}
    \label{eval:tab:ablation}
\end{table}

\parabf{Intra-stage fusion.} We compare \sysname's simulated annealing algorithm against
the greedy approach mentioned in \S\ref{sec:intra_stage_fusing:schedule}
and show the latency speedup and peak activation memory usage
relative to the serial execution of the two models with 1F1B schedule in Table~\ref{eval:tab:ablation}.
For the latency, a naive solution (denoted as 1F1B+) that does not employ a fused pipeline involves using
a smaller PP size without out-of-memory errors.
However, this approach increases the DP size and results in fewer
micro-batches allocated to each pipeline, thereby partially negating the benefits
brought by shallower pipelines. Consequently, this solution is less effective compared
to the fused pipeline schedule.
We also include a lower bound (denoted as LB) estimation calculated as follows:
we first compute the earliest possible completion time for each stage,
which consists of three parts: the earliest possible arrival time of
the first task, the total time required to process all the tasks,
and the remaining time needed for the subsequent pipeline stages of the final task;
we then take the maximum of these times across all stages as the lower bound of
the fused pipeline schedule.
Note that there may not necessarily exist a schedule that reaches this lower bound,
but we can use it as a metric to assess how close our algorithm is to the optimal
solution. Under all the settings,
our approach outperforms the baseline methods and achieves the theoretical lower bound except for the last case.

As for the memory cost, the serial 1F1B execution serves as the lower bound for the
fused schedule. Our method also greatly outperforms the greedy approach and
often reaches the lower bound. Even when it does not achieve the lower bound,
the extra overhead remains within an acceptable range.

\begin{figure*}[t!]
    \centering
    \includegraphics[width=.9\linewidth]{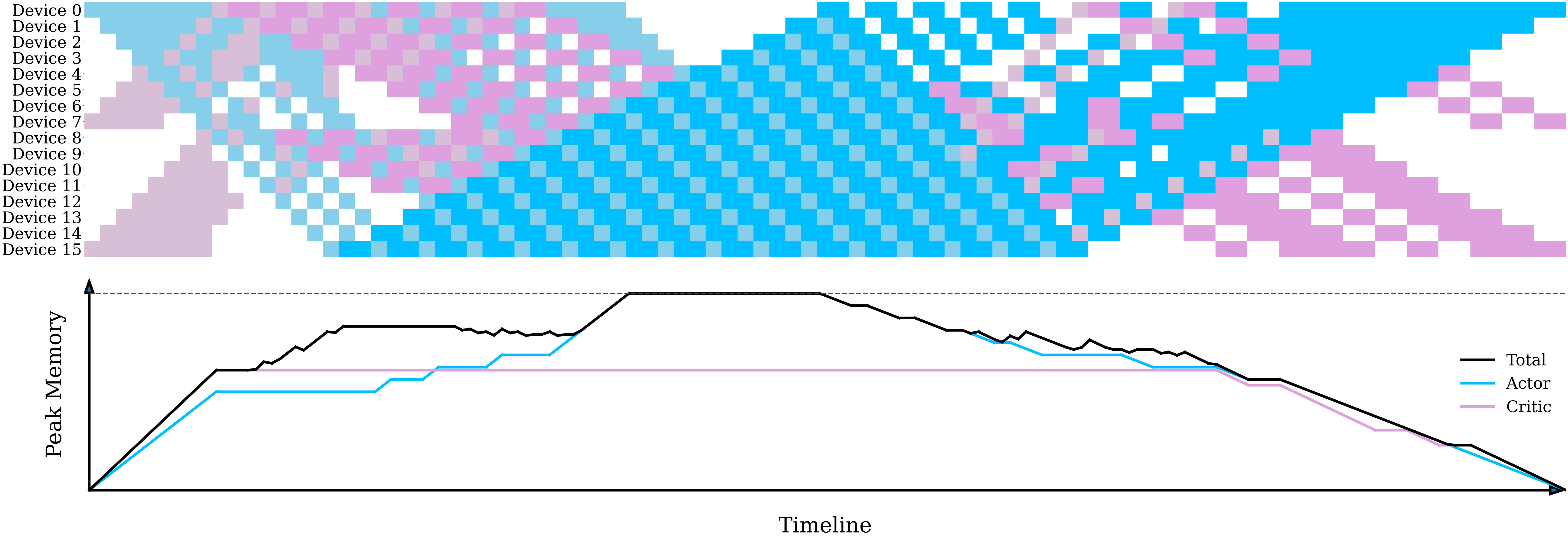}
    \vspace*{-0.1in}
    \caption{The fused pipeline schedule of 65B \texttt{Actor} and 33B \texttt{Critic} model generated by \sysname.
    The two models have 16 and 8 PP stages respectively and \#micro-batches is set equal to PP.
    Top: the GPU execution timeline. Bottom: the peak activation memory usage timeline. The red dotted horizontal line denotes
    the memory cost of serial 1F1B execution, which is the lower bound.}
    \vspace*{-0.2in}
    \label{fig:eval:case_study}
\end{figure*}

\vspace{-0.1in}
\subsection{Deep Dive}
\label{sec:eval:case_study}

In this section, we deep dive into \sysname by showing a fused pipeline schedule produced by \sysname for the
65B/33B model setting. The upper of Figure~\ref{fig:eval:case_study} shows 
the GPU execution timeline, with forward and backward tasks in light and dark colors, respectively. With intra-stage fusion,
\sysname strategically fuses one 65B model of 16 PP stages (blue grids) with two 33B models
that each have 8 PP stages (pink grids) to fill the pipeline bubbles of each other
in reversed pipeline directions.
Ultimately, the execution time of this fused pipeline schedule is the same as individually executing the 65B model
using the 1F1B schedule, which means we completely overlap the training of the
33B model, achieving the theoretical lower bound. Meanwhile, as shown at 
the bottom of Figure~\ref{fig:eval:case_study}, the peak activation
memory usage also achieves the lower bound of executing the two models
in serial with 1F1B schedule. 

\vspace{-0.1in}
\section{Related Work}
\label{sec:related}

\paraf{RLHF training systems.}
As RLHF gradually becomes the primary approach for LLM alignment,
many frameworks are specifically designed for RLHF training.
OpenRLHF~\cite{hu2024openrlhf} uses Ray~\cite{ray} to distribute models onto separate
GPUs and adopts vLLM~\cite{vllm} to accelerate the generation stage.
HybridFlow~\cite{sheng2024hybridflow} proposes a hierarchical hybrid programming model for
the RLHF dataflow and optimizes the GPU allocation and placement of each
RLHF model.
ReaLHF~\cite{mei2024realhf}
proposes to design custom parallel strategies for different
RLHF tasks to avoid GPU under-utilization.
PUZZLE~\cite{lei2024puzzle} utilizes a lightweight context-switching algorithm to reduce
the task switching overhead.
These solutions do not consider subtask-level optimization opportunities, thus
suffering from data skewness and pipeline bubble issues.

\parabf{LLM generation optimizations.}
Many systems focus on optimizing the LLM generation performance~\cite{yu2022orca, kwon2023efficient, zheng2024sglangefficientexecutionstructured, hong2023flashdecoding++, chang2024flux},
and most of them have been adopted by \sysname to accelerate the generation stage.
Besides, AlpaServe~\cite{li2023alpaserve}
multiplexes the LLM execution with colocated parallelism to
improve the throughput under the bursty workload.
FastServe~\cite{wu2023fast} proposes preemptive scheduling on iteration-level
to mitigate the head-of-line blocking caused by long-tailed samples.
Splitwise~\cite{patel2024splitwise} and DistServe~\cite{zhong2024distserve} split the prefill and decoding phases
to avoid interference between them.
These works are orthogonal to \sysname and can be integrated into \sysname.

\parabf{LLM training optimizations.}
LLM training has been studied by many works from
various aspects. In particular, \cite{huang2019gpipe,fan2021dapple,narayanan2021efficient,qi2023zero}
use pipeline parallelism to improve the training throughput and
reduce memory footprint.
Alpa~\cite{alpa} automatically generates the optimal
parallel strategy to improve training performance.
MegaScale~\cite{jiang2024megascale} provides a detailed experience in
building a production LLM training system at a large scale.
DistTrain~\cite{zhang2024disttrain} focuses on addressing the model and data heterogeneity
in multimodal LLM training.
These works are agnostic to RLHF thus do not utilizes the
characteristics to optimize the RLHF training stage.

\vspace{-0.1in}
\section{Conclusion}
\label{sec:conclusion}
We present \sysname, an efficient RLHF training system. 
\sysname views the RLHF workflow from a finer-grained subtask-level
perspective and opens up opportunities for efficient inter- and intra-stage fused execution,
mitigating data skewness and pipeline bubbles in existing systems.
The evaluation shows that \sysname can significantly
increase GPU utilization, boosting training throughput by up to $3.7\times$
compared to existing systems.

\parabf{Acknowledgments.} We thank our shepherd, Peter Pietzuch, and the
anonymous reviewers for their valuable feedback. We thank Shihong Deng, Shilei
Jiang, and Heng Wang for providing algorithmic and data support in this work. We
thank Wei Fu and Yi Wu for their assistance in running the baseline and for
their valuable feedback on the paper.

\label{lastpage}

{
\bibliographystyle{ieeetr}
\bibliography{paper}}

\end{document}